\documentclass{nle}
\usepackage[utf8]{inputenc}
\usepackage[capitalize,nameinlink]{cleveref}
\usepackage{graphicx,subfigure}
\usepackage{xcolor}
\usepackage{url}
\usepackage[usenames,dvipsnames]{pstricks}
\usepackage{epsfig}
\usepackage{pst-grad} % For gradients
\usepackage{pst-plot} % For axes
\usepackage[space]{grffile} % For spaces in paths
\usepackage{etoolbox} % For spaces in paths
\usepackage{comment}
\usepackage{multirow,graphicx}
 \usepackage{float}
\makeatletter % For spaces in paths

\title[A Structured Distributional Model]
      {A Structured Distributional Model \\ of Sentence Meaning and Processing}
\author[Chersoni et al.]
       {E.\ns C\ls h\ls e\ls r\ls s\ls o\ls n\ls i\ls ,\ns T\ls h\ls e\ns H\ls o\ls n\ls g\ns K\ls o\ls n\ls g\ns P\ls o\ls l\ls y\ls t\ls e\ls c\ls h\ls n\ls i\ls c\ns U\ls n\ls i\ls v\ls e\ls r\ls s\ls i\ls t\ls y
       \and E.\ns S\ls a\ls n\ls t\ls u\ls s\ls ,\ns M\ls a\ls s\ls s\ls a\ls c\ls h\ls u\ls s\ls e\ls t\ls t\ls s\ns I\ls n\ls s\ls t\ls i\ls t\ls u\ls t\ls e\ns o\ls f\ns T\ls e\ls c\ls h\ls n\ls o\ls l\ls o\ls g\ls y
       \and L.\ns P\ls a\ls n\ls n\ls i\ls t\ls t\ls o\ls ,\ns U\ls n\ls i\ls v\ls e\ls r\ls s\ls i\ls t\ls y\ns o\ls f\ns P\ls i\ls s\ls a
       \and
     A.\ns L\ls e\ls n\ls c\ls i\ls ,\ns U\ls n\ls i\ls v\ls e\ls r\ls s\ls i\ls t\ls y\ns o\ls f\ns P\ls i\ls s\ls a
     \and
     P.\ns B\ls l\ls a\ls c\ls h\ls e\ls ,\ns A\ls i\ls x\ls -\ls M\ls a\ls r\ls s\ls e\ls i\ls l\ls l\ls e\ns U\ls n\ls i\ls v\ls e\ls r\ls s\ls i\ls t\ls y
     \and
     C.\ns -\ls R.\ns H\ls u\ls a\ls n\ls g\ls ,\ns T\ls h\ls e\ns H\ls o\ls n\ls g\ns K\ls o\ls n\ls g\ns P\ls o\ls l\ls y\ls t\ls e\ls c\ls h\ls n\ls i\ls c\ns U\ls n\ls i\ls v\ls e\ls r\ls s\ls i\ls t\ls y
       }

\received{}

\usepackage{nle-chicago-patch}
\bibpunct{(}{)}{,}{a}{}{;}

\begin{document}

\label{firstpage}

\maketitle

\begin{abstract}
Most compositional distributional semantic models represent sentence meaning with a single vector.
In this paper, we propose a Structured Distributional Model (SDM) that combines word embeddings with formal semantics and is based on the assumption that sentences represent events and situations. The semantic representation of a sentence is a formal structure derived from Discourse Representation Theory and containing distributional vectors. This structure is dynamically and incrementally built by integrating knowledge about events and their typical participants, as they are activated by lexical items. Event knowledge is modeled as a graph extracted from parsed corpora and encoding roles and relationships between participants that are represented as distributional vectors. SDM is grounded on extensive psycholinguistic research showing that generalized knowledge about events stored in semantic memory plays a key role in sentence comprehension.
We evaluate SDM on two recently introduced compositionality datasets, and our results show that combining a simple compositional model with event knowledge constantly improves performances, even with different types of word embeddings.
\end{abstract}

\setbox0=\hbox{\cite{kiros2015skip}}
\setbox0=\hbox{\cite{hill2016learning}}
\setbox0=\hbox{\cite{baroni2013frege}}
\setbox0=\hbox{\cite{paperno2014practical}}
\setbox0=\hbox{\cite{rimell2016relpron}}
\setbox0=\hbox{\cite{mikolov2013distributed}}
\setbox0=\hbox{\cite{vassallo2018event}}
\setbox0=\hbox{\cite{mcrae2005basis}}
\setbox0=\hbox{\cite{mcrae1998modeling}}
\setbox0=\hbox{\cite{hare2009activating}}
\setbox0=\hbox{\cite{bicknell2010effects}}
\setbox0=\hbox{\cite{metusalem2012generalized}}
\setbox0=\hbox{\cite{matsuki2011event}}
\setbox0=\hbox{\cite{Ferretti2001516}}
\setbox0=\hbox{\cite{Erk2010AFC}}
\setbox0=\hbox{\cite{Erk2007ASS}}
\setbox0=\hbox{\cite{Sayeed2015AnEO}}
\setbox0=\hbox{\cite{Tilk2016EventPM}}
\setbox0=\hbox{\cite{Greenberg2015ImprovingUV}}
\setbox0=\hbox{\cite{zaremba2014recurrent}}
\setbox0=\hbox{\cite{santus2017measuring}}
\setbox0=\hbox{\cite{chersoni2017logical}}
\setbox0=\hbox{\cite{zarcone2012modeling}}
\setbox0=\hbox{\cite{kruszewski2015jointly}}
\setbox0=\hbox{\cite{hong2018learning}}
\setbox0=\hbox{\cite{chersoni2016representing}}
\setbox0=\hbox{\cite{manning2014stanford}}
\setbox0=\hbox{\cite{Baroni2009}}
\setbox0=\hbox{\cite{levy2015improving}}
\setbox0=\hbox{\cite{lapesa2017large}}
\setbox0=\hbox{\cite{chersoni2017structure}}
\setbox0=\hbox{\cite{Sayeed2016ThematicFE}}
\setbox0=\hbox{\cite{baroni2014don}}
\setbox0=\hbox{\cite{Beltagy:etal:2016}}
\setbox0=\hbox{\cite{palangi2018question}}
\setbox0=\hbox{\cite{conneau2018you}}
\setbox0=\hbox{\cite{zhu2018exploring}}
\setbox0=\hbox{\cite{ettinger2016probing}}
\setbox0=\hbox{\cite{adi2016fine}}
\setbox0=\hbox{\cite{bar2007units}}
\setbox0=\hbox{\cite{meltzer2015brain}}
\setbox0=\hbox{\cite{williams2017early}}
\setbox0=\hbox{\cite{Arora:etal:2017}}
\setbox0=\hbox{\cite{Tian:etal:2017}}

\section{Sentence Meaning in Vector Spaces}
\label{sec:intro}

While for decades sentence meaning has been represented in terms of complex formal structures, the most recent trend in computational semantics is to model semantic representations with dense distributional vectors (aka \emph{embeddings}). As a matter of fact, distributional semantics has become one of the most influential approaches to lexical meaning, because of the important theoretical and computational advantages of representing words with continuous vectors, such as automatically learning lexical representations from natural language corpora and multimodal data, assessing semantic similarity in terms of the distance between the vectors, and dealing with the inherently gradient and fuzzy nature of meaning \citep{Erk:2012,Lenci:2018a}.

Over the years, intense research has tried to address the question of how to project the strengths of vector models of meaning beyond word level, to phrases and sentences. The mainstream approach in distributional semantics assumes the representation of sentence meaning to be a vector, exactly like lexical items. 
Early approaches simply used pointwise vector operations (such as addition or multiplication) to combine word vectors to form phrase or sentence vectors \citep{mitchell2010composition}, and in several tasks they still represent a non-trivial baseline to beat \citep{rimell2016relpron}.
More recent contributions can be essentially divided into two separate trends. The former attempts to model `Fregean compositionality' in vector space, and aimes at finding progressively more sophisticated compositional operations to derive sentence representations from the vectors of the words composing them \citep{baroni2013frege,paperno2014practical}. 
In the latter trend, dense vectors for sentences are learned as a whole, in a similar way to neural word embeddings \citep{mikolov2013distributed,levy2014neural}: for example, the encoder-decoder models of works like \citet{kiros2015skip} and \citet{hill2016learning} are trained to predict, given a sentence vector, the vectors of the surrounding sentences.

Representing sentences with vectors appears to be unrivaled from the applicative point of view, and has indeed important advantages such as the possibility of measuring similarity between sentences with their embeddings, as is customary at the lexical level, which is then exploited in tasks like automatic paraphrasing and captioning, question-answering, etc. Recently, probing tasks have been proposed to test what kind of syntactic and semantic information is encoded in sentence embeddings \citep{ettinger2016probing,adi2016fine,conneau2018you,zhu2018exploring}. In particular, \cite{zhu2018exploring} show that current models are not able to discriminate between different syntactic realization of semantic roles, and fail to recognize that \emph{Lilly loves Imogen} is more similar to 
its passive counterpart than to \emph{Imogen loves Lilly}. Moreover, it is difficult to recover information about the component words from sentence embeddings \citep{adi2016fine,conneau2018you}. 
The semantic representations built with tensor product in the question-answering system by \citet{palangi2018question} have been claimed to be grammatically interpretable as well.
However, the complexity of the semantic information brought by sentences and the difficulty to interpret the embeddings raise doubts about the general theoretical and empirical validity of the ``sentence-meaning-as-vector'' approach.

In this paper, we propose a \textbf{Structured Distributional Model} (SDM) of sentence meaning that combines word embeddings with formal semantics and is based on the assumption that sentences represent events and situations. These are regarded as inherently complex semantic objects, involving multiple entities that interact with different roles (e.g., agents, patients, locations etc.). The semantic representation of a sentence is a formal structure inspired by Discourse Representation Theory (DRT) \citep{Kamp:2013} and containing distributional vectors. This structure is dynamically and incrementally built by integrating knowledge about events and their typical participants, as they are activated by lexical items. Event knowledge is modeled as a graph extracted from parsed corpora and encoding roles and relationships between participants that are represented as distributional vectors.  The semantic representations of SDM retain the advantages of embeddings (e.g., learnability, gradability, etc.), but also contain directly interpretable formal structures, differently from classical vector-based approaches. 

SDM is grounded on extensive psycholinguistic research showing that generalized knowledge about events stored in semantic memory plays a key role in sentence comprehension \citep{mcrae2009people}. On the other hand, it is also close to recent attempts to look for a ``division of labour'' between formal and vector semantics, representing sentences with logical forms enriched with distributional representations of lexical items \citep{Beltagy:etal:2016,Boleda:Herbelot:2016,McNally:2017}. Like SDM, \cite{McNally:Boleda:2017} propose to introduce embeddings within DRT semantic representations. At the same time, differently from these other approaches, SDM consists of formal structures that integrate word embeddings with a distributional representation of activated event knowledge, which is then dynamically integrated during semantic composition.

The contribution of this paper is twofold. First, we introduce SDM as a cognitively-inspired distributional model of sentence meaning based on a structured formalization of semantic representations and contextual event knowledge (Section \ref{sec:Model}).  Secondly, we show that the event knowledge used by SDM in the construction of sentence meaning representations leads to improvements over other state-of-the-art models in compositionality tasks. In Section \ref{sec:Exp}, SDM is tested on two different benchmarks: the first is RELPRON \citep{rimell2016relpron}, a popular dataset for the similarity estimation between compositional distributional representations; the second is  DTFit \citep{vassallo2018event}, a dataset created to model an important aspect of sentence meaning, that is the typicality of the described event or situation, which has been shown to have important processing consequences for language comprehension.

\section{Dynamic Composition with Embeddings and Event Knowledge}
\label{sec:Model}

SDM rests on the assumption that natural language comprehension involves the \emph{dynamic construction of semantic representations, as mental characterization of the events or situations described in sentences}. We use the term `dynamic' in the sense of dynamic semantic frameworks like DRT, to refer to a bidirectional relationship between linguistic meaning and context \citep[see also][]{Heim:1983}:

\begin{quote}
\noindent{}The meaning of an expression depends on the context in which it is used, and its content is itself defined as a \emph{context-change potential}, which affects and determines the interpretation of the following expressions.
\end{quote}

\noindent{}The content of an expression $E$ used in a context $C$ depends on $C$, but -- once the content has been determined -- it will contribute to update $C$ to a new context $C'$, which will help fixing the content of the next expression. 
Similarly to DRT, SDM integrates word embeddings in a dynamic process to construct the semantic representations of sentences. Contextual knowledge is represented in distributional terms and affects the interpretation of following expressions, which in turn cue new information that updates the current context.\footnote{An early work on a distributional model of lexical expectations in context is \citet{washtell2010expectation}, but its focus was more on word sense disambiguation than on representing sentence meaning.}

Context is a highly multifaceted notion that includes several types of factors guiding and influencing language comprehension: information about the communicative settings, preceding discourse, general presuppositions and knowledge about the world, etc. In DRT, \cite{Kamp:2016} has introduced the notion of \emph{articulated context} to model different sources of contextual information that intervene in the dynamic construction of semantic representations. In this paper, we focus on the contribution of a specific type of contextual information, which we refer to as \emph{Generalized Event Knowledge} (\textsc{gek}). This is knowledge about events and situations that we have experienced under different modalities, including the linguistic input \citep{mcrae2009people}, and is generalized because it contains information about prototypical event structures.

In linguistics, the Generative Lexicon theory \citep{Pustejovsky:1995} argues that the lexical entries of nouns also contain information about events that are crucial to define their meaning (e.g., \emph{read} for \emph{book}).
Psycholinguistic studies in the last two decades have brought extensive evidence that the array of event knowledge activated during sentence processing is extremely rich: verbs (e.g. \textit{arrest}) activate expectations about typical arguments (e.g. \textit{cop, thief}) and vice versa \citep{mcrae1998modeling,Ferretti2001516,mcrae2005basis}, and similarly nouns activate other nouns typically co-occurring as participants in the same events (\textit{key, door}) \citep{hare2009activating}.
The influence of argument structure relations on how words are neurally processed is also an important field of study in cognitive neuroscience \citep{thompson2014neurocognitive,meltzer2015brain,williams2017early}.

Stored event knowledge has relevant processing consequences. Neurocognitive research showed that the brain is constantly engaged in making predictions to anticipate future events \citep{Bar:2009,Clark:2013}. Language comprehension, in turn, has been characterized as a largely predictive process \citep{Kuperberg:Jaeger:2015}. Predictions are memory-based, and experiences about events and their participants are used to generate expectations about the upcoming linguistic input, thereby minimizing the processing effort \citep{elman20145,mcrae2009people}. For instance, argument combinations that are more `coherent' with the event scenarios activated by the previous words are read faster in self-paced reading tasks and elicited smaller N400 amplitudes in ERP experiments \citep{bicknell2010effects,matsuki2011event,Paczynski:Kuperberg:2012,metusalem2012generalized}.\footnote{Event-related potentials are the electrophysiological response of the brain to a stimulus. In the sentence processing literature, the ERPs are recorded for each stimulus word and the N400, one of the most studied ones, is a negative-going deflection appearing 400ms after the presentation of the word. 
A common interpretation of the N400 assumes that the wave amplitude is proportional to the difficulty of semantic unification \citep{baggio2011balance}.}

\cite{elman2009meaning,elman20145} has proposed a general interpretation of these experimental results in the light of the Words-as-Cues framework. According to this theory, words are arranged in the mental lexicon as a sort of network of mutual expectations, and listeners rely on pre-stored representations of events and common situations to try to identify the one that a speaker is more likely to communicate. As new input words are processed, they are quickly integrated in a data structure containing a dynamic representation of the sentence content, until some events are recognized as the `best candidates' for explaining the cues (i.e., the words) observed in the linguistic input.  It is important to stress that, in such a view, the meaning of complex units such as phrases and sentences is not always built by composing lexical meanings, as the representation of typical events might be already stored and retrieved as a whole in semantic memory. Participants often occurring together become active when the representation of one of them is activated (see also Bar et al., 2007 on the relation between associative processing and predictions).

SDM aims at integrating the core aspects of dynamic formal semantics and the evidence on the role of event knowledge for language processing into a general model for compositional semantic representations that relies on two major assumptions:

\begin{itemize}
\item lexical items are represented as embeddings within a network of relations encoding knowledge about events and typical participants, which corresponds to what we have termed above \textsc{gek};
\item the \emph{semantic representation} (\textsc{sr}) of a sentence (or even larger stretches of linguistic input, such as discourse) is a formal structure that dynamically combines the information cued by lexical items.
\end{itemize}

\noindent{}Like in \citet{chersoni2017logical}, the model is inspired by Memory, Unification and Control (MUC), proposed by Hagoort \citep{Hagoort:2013,Hagoort:2016} as a general model for the neurobiology of language. MUC incorporates three main functional components: i.) \emph{Memory} corresponds to knowledge stored in long-term memory; ii.) \emph{Unification} refers to the process of combining the units stored in \emph{Memory} to create larger structures, with contributions from the context; and iii.) \emph{Control} is responsible for relating language to joint action and social interaction. Similarly, our model
distinguishes between a component storing event knowledge, in the form of a \textbf{Distributional Event Graph} (\textsc{deg}, Section \ref{sec:DEG}), and a \textbf{meaning composition function} that integrates information activated from lexical items and incrementally builds the \textsc{sr} (Section \ref{sec:MCF}).

\subsection{The Distributional Event Graph}
\label{sec:DEG}

The Distributional Event Graph represents the event knowledge stored in long-term memory with information extracted from parsed corpora. We assume a very broad notion of \emph{event}, as an $n$-ary relation between entities. Accordingly, an event can be a complex situation involving multiple participants, such as \emph{The student reads a book in the library}, but also the association between an entity and a property expressed by the noun phrase \emph{heavy book}. This notion of event corresponds to what psychologist call \emph{situation knowledge} or \emph{thematic associations} \citep{Binder:2016}. As \cite{mcrae2009people} argue, \textsc{gek} is acquired from both sensorimotor experience (e.g., watching or playing football matches) and linguistic experience (e.g., reading about football matches). \textsc{deg} can thus be regarded as a model of the \textsc{gek} derived from the linguistic input.

\begin{figure}
  \includegraphics[width=0.7\textwidth]{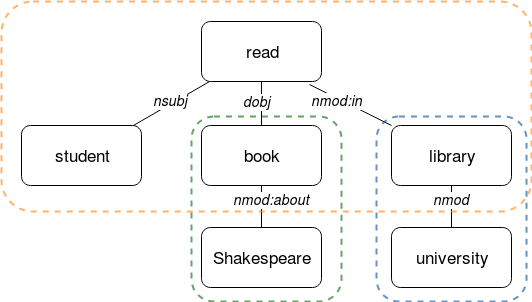}
  \caption{Reduced version of the parsing for the sentence \textit{The student is reading the book about Shakespeare in the university library}. Three events are identified, each represented with a dotted box.}
  \label{deps}
\end{figure}

Events are extracted from parsed sentences, using syntactic relations as an approximation of deeper semantic roles (e.g., the subject relation for the agent, the direct object relation for the patient, etc.). In the present paper, we use dependency parses, as it is customary in distributional semantics, but nothing in SDM hinges on the choice of the syntactic representation. Given a verb or a noun head, all its syntactic dependents are grouped together.\footnote{The extracted graphs are similar to the syntactic joint contexts for verb representation that were proposed by \citet{chersoni2016representing}.} More schematic events are also generated by abstracting from one or more event participants for every recorded instance. Since we expect each participant to be able to trigger the event and consequently any of the other participants, a relation can be created and added to the graph from every subset of each group extracted from a sentence (cf. Figure \ref{deps}).

The resulting \textsc{deg} structure is a \textit{weighted hypergraph}, as it contains weighted relations holding between nodes pairs, and a \textit{labeled multigraph}, since the edges are labeled in order to represent specific syntactic relations. The weights $\sigma$ are derived from co-occurrence statistics and measure the association strengths between event nodes. They are intended as salience scores that identify the most prototypical events associated with an entity (e.g., the typical actions performed by a student).
Crucially, the graph nodes are represented as word embeddings. Thus, given a lexical cue $w$, the information in \textsc{deg} can be activated along two dimensions during processing (cf. Table \ref{tab:DEG}):

\begin{enumerate}
\item by retrieving the most similar nodes to $w$ (the paradigmatic neighbors), on the basis of their cosine similarity between their vectors and the vector of $w$;
\item by retrieving the closest associates of $w$ (the syntagmatic neighbors), using the edge weights.
\end{enumerate}

\begin{figure}
  \includegraphics[scale=0.63]{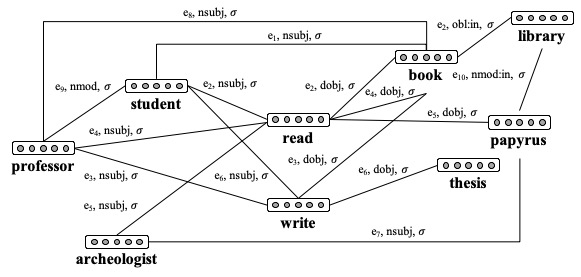}
  \caption{Toy sample of \textsc{deg} showing several instances of events, each represented by a sequence of co-indexed $e$. The $\sigma$ are the event salience weights.}
  \label{fig:DEG}
\end{figure}

\noindent{}Figure \ref{fig:DEG} shows a toy example of \textsc{deg}. The little boxes with circles in them represent the embedding associated with each node. Edges are labeled with syntactic relations (as a surface approximation of event roles) and weighted with salience scores $\sigma$. Each event is a set of co-indexed edges. For example, $e_2$ corresponds to the event of students reading books in libraries, while $e_1$ represents a schematic event of students performing some generic action on books (e.g., reading, consulting, studying, etc.).

\begin{table}
\begin{tabular}{cc}
\textbf{Paradigmatic Neighbors}  & \textbf{Syntagmatic Neighbors}  \\ \hline
essay, story, novel, author,
biography & publish, write, read, child, series \\
\hline
\end{tabular}
\caption{The five nearest paradigmatic and syntagmatic neighbors for the lexical item \textnormal{book}, extracted from \textsc{deg}.}
\label{tab:DEG}
\end{table}

\subsection{The Meaning Composition Function}
\label{sec:MCF}

We assume that during sentence comprehension lexical items activate fragments of event knowledge stored in \textsc{deg} (like in Elman's Words-as-Cues model), which are then dynamically integrated in a semantic representation \textsc{sr}. This is a formal structure directly inspired by DRT and consisting of three different yet interacting information tiers:

\begin{enumerate}[i]
\item \textit{universe} (\textsc{U}) - this tier, which we do not discuss further in the present paper, includes the entities mentioned in the sentence (corresponding to the \emph{discourse referents} in DRT). They are typically introduced by noun phrases and provide the targets of anaphoric links;
\item \textit{linguistic conditions} (\textsc{lc}) - a context-independent tier of meaning that accumulates the embeddings associated with the lexical items. This corresponds to the conditions that in DRT content words add to the discourse referents. The crucial difference is that now such conditions are embeddings;
\item \textit{active context} (\textsc{ac}) - similarly to the notion of \emph{articulated context} in \cite{Kamp:2016}, this component consists of several types of contextual information available during sentence processing or activated by lexical items (e.g., information from the current communication setting, general world knowledge, etc.). More specifically, we assume that \textsc{ac} contains the embeddings activated from \textsc{deg} by the single lexemes (or by other contextual elements) and integrated into a semantically coherent structure contributing to the sentence interpretation.
\end{enumerate}

\begin{figure}
  \includegraphics[scale=0.45]{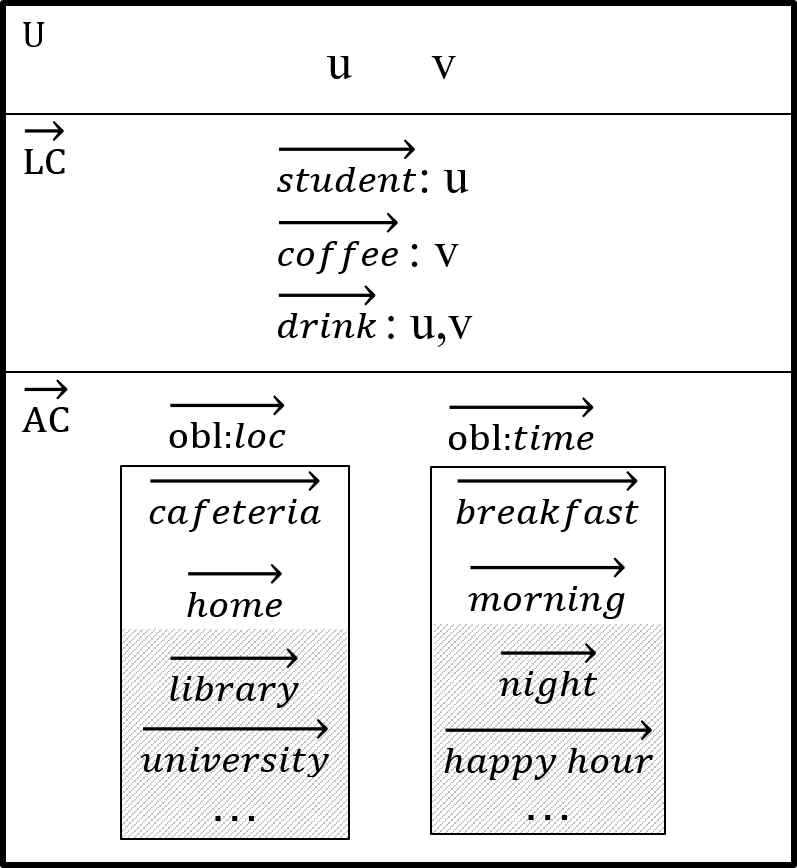}
    % the space for the artwork
  \caption{Sample \textsc{sr} for the sentence \emph{The student drinks the coffee}. The sentence activates typical locations and times in which the event could take place.}
  \label{fig:SR1}
\end{figure}

\noindent{}Figure \ref{fig:SR1} shows an example of \textsc{sr} built from the sentence \emph{The student drinks the coffee} (ignoring the specific contribution of determiners and tense). The universe \textsc{U} contains the discourse referents introduced by the noun phrases, while \textsc{lc} includes the embeddings of the lexical items in the sentence, each linked to the relevant referent (e.g., $\overrightarrow{student}:u$ means that the embedding introduced by \emph{student} is linked to the discourse referent $u$). \textsc{ac} consists of the embeddings activated from \textsc{deg} and ranked by their salience with respect to the current content in the \textsc{sr}. The elements in \textsc{ac} are grouped by their syntactic relation in \textsc{deg}, which again we regard here just as a surface approximation of their semantic role (e.g., the items listed under ``obl:\emph{loc}'' are a set of possible locations of the event expressed by the sentence). \textsc{ac} makes it possible to enrich the semantic content of the sentence with contextual information, predict other elements of the event, and generate expectations about incoming input. For instance, given the \textsc{ac} in Figure \ref{fig:SR1}, we can predict that the student is most likely to be drinking a coffee at the cafeteria and that he/she is drinking it for breakfast or in the morning. The ranking of each element in \textsc{ac} depends on two factors: i.) its degree of activation by the lexical items, ii.) its overall coherence with respect to the information already available in the \textsc{ac}.

\begin{figure}[t]
  \includegraphics[scale=0.40]{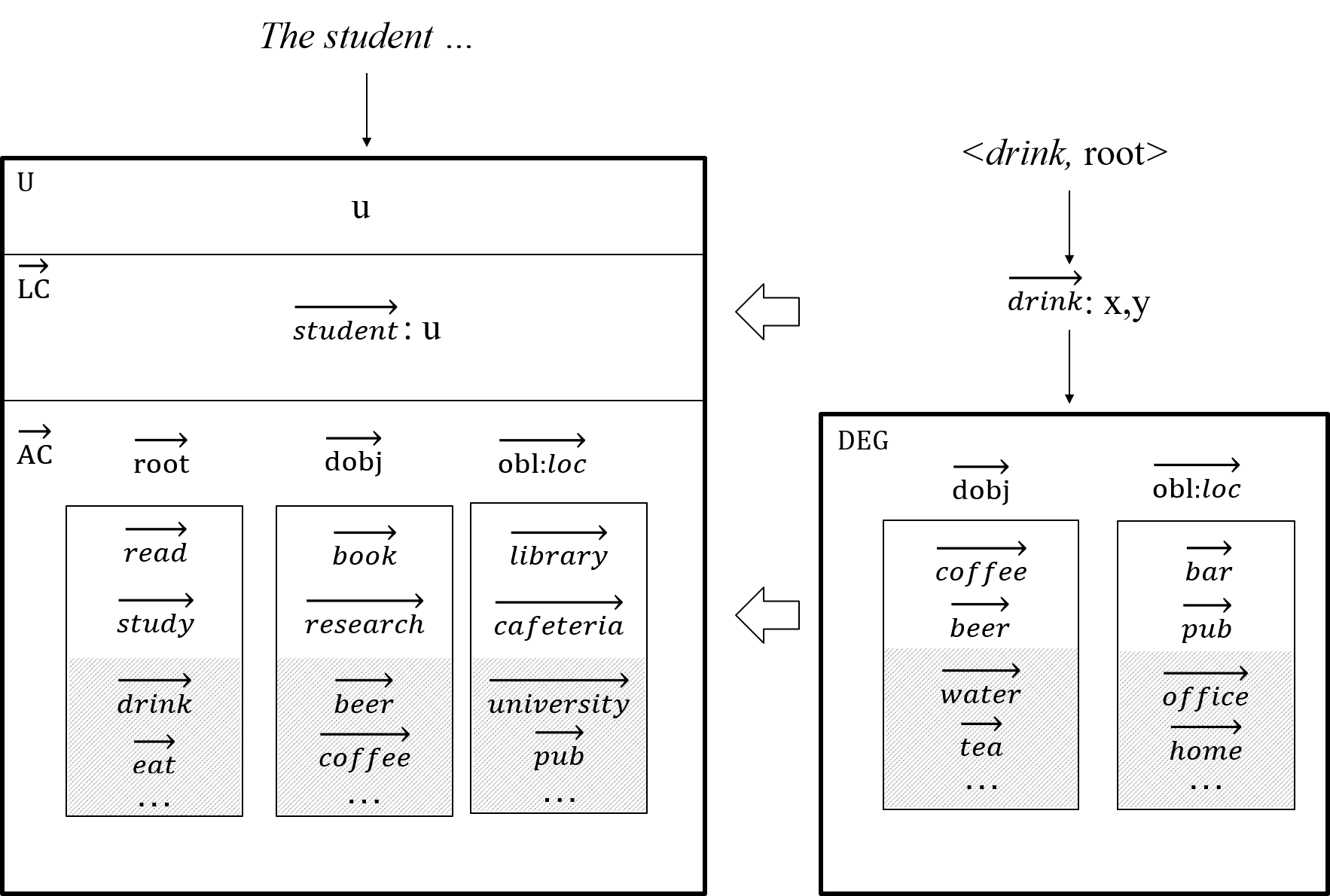}
    % the space for the artwork
  \caption{On the left, the \textsc{sr} for \emph{The student}. On the right, the embedding and \textsc{deg} portion activated by the verb \emph{drink}.}
  \label{fig:SR2}
\end{figure}

\begin{figure}[t]
  \includegraphics[scale=0.40]{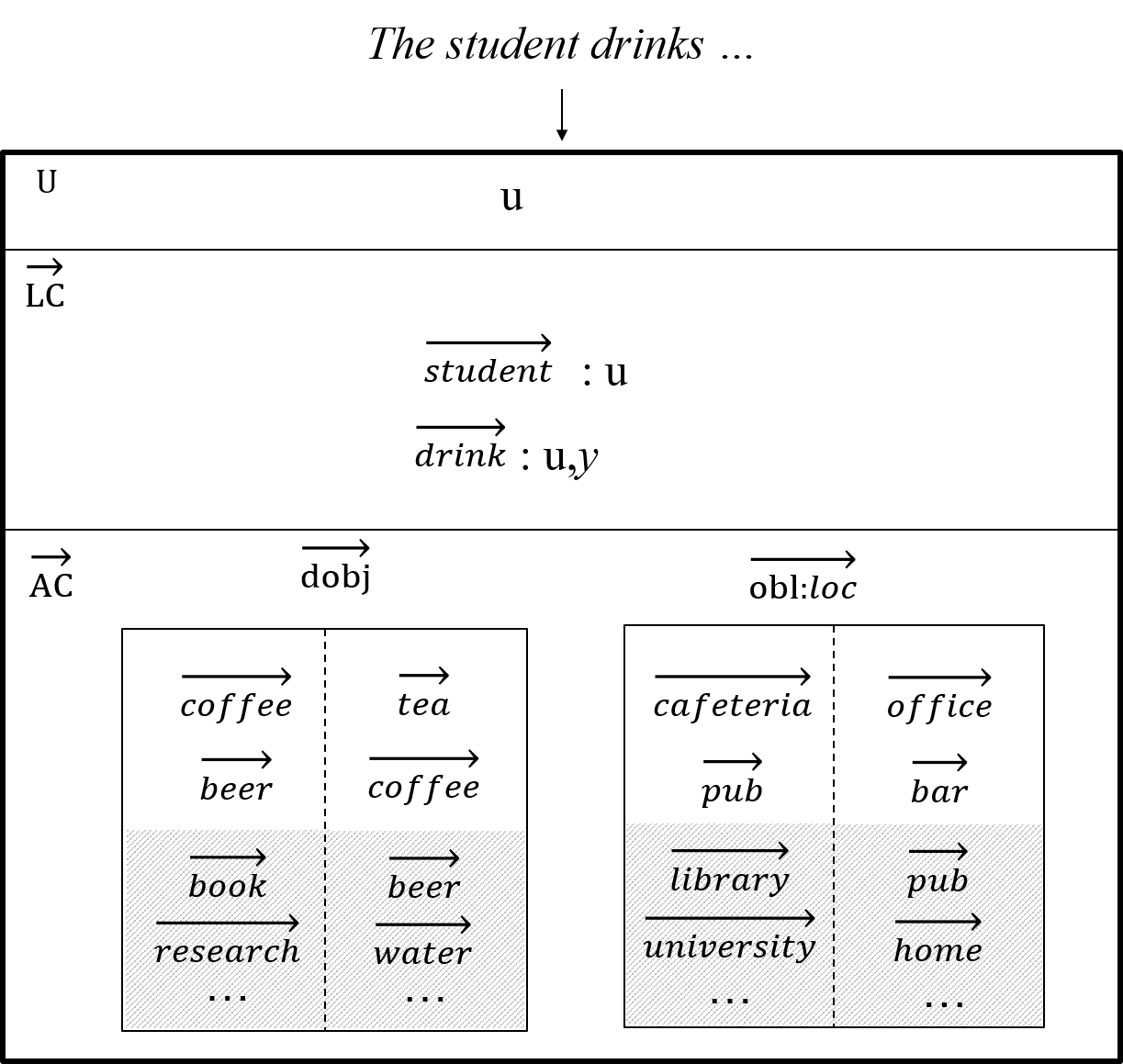}
    % the space for the artwork
  \caption{The original semantic representation \textsc{sr} for \emph{The student $\dots$} is updated with the information activated by the verb, producing the \textsc{sr} for \emph{The student drinks $\dots$} The new event knowledge is re-ranked with respect to the previous content of \textsc{ac}.}
  \label{fig:SR3}
\end{figure}

A crucial feature of each \textsc{sr} is that \textsc{lc} and \textsc{ac} are also represented with vectors that are incrementally updated with the information activated by lexical items. Let \textsc{sr}$_{i-1}$ be the semantic representation built for the linguistic input $w_1,\dots, w_{i-1}$. When we process a new pair $\langle w_i,r_i \rangle$ with a lexeme $w_i$ and syntactic role $r_i$:
\begin{enumerate}[i.)]
\item \textsc{lc} in \textsc{sr}$_{i-1}$ is updated with the embedding $\overrightarrow{w_i}$;
\item \textsc{ac} in \textsc{sr}$_{i-1}$ is updated with the embeddings of the syntagmatic neighbors of $w_i$ extracted from \textsc{deg}.
\end{enumerate}
Figures \ref{fig:SR2} and \ref{fig:SR3} exemplify the update of the \textsc{sr} for the subject \emph{The student} with the information activated by the verb \emph{drink}. The update process is defined as follows:

\begin{enumerate}
\item \textsc{lc} is represented with the vector $\overrightarrow{LC}$ obtained from the linear combination of the embeddings of the words contained in the sentence. Therefore, when $\langle w_i,r_i\rangle$ is processed, the embedding $\overrightarrow{w_i}$ is simply added to $\overrightarrow{LC}$;\footnote{At the same time, the embedding is linked either to a new discourse referent added to \textsc{U}, or to an already available one.}
\item for each syntactic role $r_i$, \textsc{ac} contains a set of ranked lists (one for each processed pair) of embeddings corresponding to the most likely words expected to fill that role. For instance, the \textsc{ac} for the fragment \emph{The student} in Figure \ref{fig:SR2} contains a list of the embeddings of the most expected direct objects associated with \emph{student}, a list of the embeddings of the most expected locations, etc.  Each list of expected role fillers is itself represented with the weighted centroid vector (e.g., $\overrightarrow{dobj}$) of their $k$ most prominent items (with $k$ a model hyperparameter). For instance, setting $k=2$, the $\overrightarrow{dobj}$ centroid in the \textsc{ac} in figure \ref{fig:SR2} is built just from $\overrightarrow{book}$ and $\overrightarrow{research}$; less salient elements (the gray areas in Figures \ref{fig:SR1}, \ref{fig:SR2} and \ref{fig:SR3}) are kept in the list of likely direct objects, but at this stage do not contribute to the centroid representing the expected fillers for that role. \textsc{ac} is then updated with the \textsc{deg} fragment activated by the new lexeme $w_i$ (e.g., the verb \textit{drink}):
\begin{itemize}
\item the event knowledge activated by $w_i$ for a given role $r_i$  is ranked according to cosine similarity with the vector $\overrightarrow{r_i}$ available in \textsc{ac}: in our example, the direct objects activated by the verb \textit{drink} (e.g., $\overrightarrow{beer}$, $\overrightarrow{coffee}$, etc.) are ranked according to their cosine similarity to the $\overrightarrow{dobj}$ vector of the \textsc{ac};
\item the ranking process works also in the opposite direction: the newly retrieved information is used to update the centroids in \textsc{ac}. For example, the direct objects activated by the verb \textit{drink} are aggregated into centroids and the corresponding weighted lists in \textsc{ac} are re-ranked according to the cosine similarity with the new centroids, in order to maximize the semantic coherence of the representation. At this point, $\overrightarrow{book}$ and $\overrightarrow{research}$, which are not as salient as $\overrightarrow{coffee}$ and $\overrightarrow{beer}$ in the \textit{drinking} context, are downgraded in the ranked list and are therefore less likely to become part of the $\overrightarrow{dobj}$ centroid at the next step.
\end{itemize}
The newly retrieved information is now added to the \textsc{ac}: as shown in Figure \ref{fig:SR3}, once the pair $\langle drink, root \rangle$ has been fully processed, the \textsc{ac} contains two ranked lists for the \textit{dobj} role and two ranked lists for the \textit{obl:loc} role, the top \textit{k} elements of each list will be part of the centroid for their relation in the next step. Finally, the whole \textsc{ac} is represented with the centroid vector $\overrightarrow{AC}$ built out of the role vectors $\overrightarrow{r_1},\dots,\overrightarrow{r_n}$ available in \textsc{ac}. The vector $\overrightarrow{AC}$ encodes the integrated event knowledge activated by the linguistic input.
\end{enumerate}

As an example of \textsc{gek} re-ranking, assume that after processing the subject noun phrase \emph{The student}, the \textsc{ac} of the corresponding \textsc{sr} predicts that the most expected verbs are \emph{read, study, drink}, etc., the most expected associated direct objects are \emph{book, research, beer}, etc., and the most expected locations are \emph{library, cafeteria, university,} etc. (Figure \ref{fig:SR2}).  When the main verb \emph{drink} is processed, the corresponding role list is removed by the \textsc{ac}, because that syntactic slot is now overtly filled by this lexeme, whose embedding is then added to the \textsc{lc}. The verb \emph{drink} cues its own event knowledge, for instance that the most typical objects of drinking are \emph{tea, coffee, beer}, etc., and the most typical locations are \emph{cafeteria, pub, bar}, etc. The information cued by \emph{drink} is re-ranked to promote those items that are most compatible and coherent with the current content of \textsc{ac} (i.e., direct objects and locations that are likely to interact with students). Analogously, the information in the \textsc{ac} is re-ranked to make it more compatible with the \textsc{gek} cued by \emph{drink} (e.g., the salience of \emph{book} and \emph{research} gets decreased, because they are not similar to the typical direct objects and locations of \emph{drink}). The output of the \textsc{sr} update is shown in Figure \ref{fig:SR3}, whose \textsc{ac} now contains the \textsc{gek} associated with an event of drinking by a student.

A crucial feature of \textsc{sr} is that it is a much richer representation than the bare linguistic input: the overtly realized arguments in fact activate a broader array of roles than the ones actually appearing in the sentence. As an example of how these unexpressed arguments contribute to the semantic representation of the event, consider a situation in which three different sentences are represented by means of \textsc{ac}, namely \textit{The student writes the thesis}, \textit{The headmaster writes the review} and \textit{The teacher writes the assignment}. Although \textit{teacher} could be judged as closer to \textit{headmaster} than to \textit{student}, and \textit{thesis} as closer to \textit{assignment} than to \textit{review}, taking into account also the typical locations (e.g., a \textit{library} for the first two sentences, a \textit{classroom} for the last one) and writing supports (e.g., a \textit{laptop} in the first two cases, a \textit{blackboard} in the last one) would lead to the first two events being judged as the most similar ones.

In the case of unexpected continuations, the \textsc{ac} will be updated with the new information, though in this case the re-ranking process would probably not change the \textsc{gek} prominence. Consider the case of an input fragment like \textit{The student plows...}: \textit{student} activates event knowledge as it is shown in Figure \ref{fig:SR1}, but the verb does not belong to the set of expected events given \emph{student}. The verb triggers different direct objects from those already in the \textsc{ac} (e.g., typical objects of \textit{plow} such as \textit{furrow}, \emph{field}, etc.). Since the similarity of their centroid with the elements of the direct object list in the \textsc{ac} will be very low, the relative ordering of the ranked list will roughly stay the same, and direct objects pertaining to the plowing situation will coexist with direct objects triggered by \textit{student}.
Depending on the continuation of the sentence, then, the elements triggered by \textit{plow} might gain centrality in the representation or remain peripheral.

It is worth noting that the incremental process of the \textsc{sr} update is consistent with the main principles of formal dynamic semantic frameworks like DRT. As we said above, dynamics semantics assumes the meaning of an expression to be a context-change potential that affects the interpretation of the following expressions. Similarly, in our distributional model of sentence representation the \textsc{ac} in  \textsc{sr}$_{i-1}$ affects the interpretation of the incoming input $w_i$, via the \textsc{gek} re-ranking process.\footnote{For a more comprehensive analysis of the relationship between distributional semantics and dynamics semantics, see \cite{Lenci:2018}.}

\section{Experiments}
\label{sec:Exp}

\subsection{Datasets and Tasks}

Our goal is to test SDM in compositionality-related tasks, with a particular focus on the contribution of event knowledge. For the present study, we selected two different datasets: the development set of the RELPRON dataset \citep{rimell2016relpron}\footnote{We used the development set of RELPRON in order to compare our results with those published by \citet{rimell2016relpron}.} and the DTFit dataset \citep{vassallo2018event}.

\begin{figure}[t]
  \includegraphics[width=0.5\textwidth]{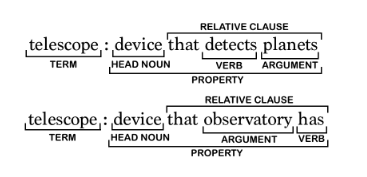}
  \caption{Image from \cite{rimell2016relpron}, showing the terminology for terms and properties in RELPRON: subject relative clause top, object relative clause bottom.}
  \label{fig:relpron-exe}
\end{figure}

\textbf{RELPRON} consists of 518 target-property pairs, where the target is a noun labeled with a syntactic function (either subject or direct object) and the property is a subject or object relative clause providing the definition of the target (Figure \ref{fig:relpron-exe}). Given a model, we produce a compositional representation for each of the properties. In each definition, the \textit{verb}, the \textit{head noun} and the \textit{argument} are composed to obtain a representation of the property. Following the original evaluation in \cite{rimell2016relpron}, we tested six different combinations for each composition model: the verb only, the argument only, the head noun and the verb, the head noun and the argument, the verb and the argument and all three of them. 
For each target, the 518 composed vectors are ranked according to their cosine similarity to the target. Like \citet{rimell2016relpron}, we use Mean Average Precision (henceforth MAP) to evaluate our models on RELPRON. Formally, MAP is defined as 
\begin{equation}
    MAP = \frac{1}{N}\sum_{i=1}^{N}AP(t_i)
\end{equation}
where $N$ is the number of terms in RELPRON, and $AP(t)$ is the Average Precision for
term $t$, defined as:
\begin{equation}
    AP(t) = \frac{1}{P_t}\sum_{k=1}^{M}Prec(k) \times rel(k)
\end{equation}
where $P_t$ is the number of correct properties for term $t$ in the dataset, $M$ is the total
number of properties in the dataset, $Prec(k)$ is the precision at rank $k$, and $rel(k)$ is a function equal to one if the property at rank $k$ is a correct property for $t$, and zero otherwise. Intuitively, $AP(t)$ will be $1$ if, for the term $t$, all the correct properties associated to the term are ranked in the top positions, and the value becomes lower when the correct items are ranked farther from the head of the list.

Our second evaluation dataset, \textbf{DTFit}, has been introduced with the goal of building a new gold standard for the \emph{thematic fit} estimation task \citep{vassallo2018event}. Thematic fit is a psycholinguistic notion similar to selectional preferences, the main difference being that the latter involve the satisfaction of constraints on discrete semantic features of the arguments, while thematic fit is a continuous value expressing the degree of compatibility between an argument and a semantic role \citep{mcrae1998modeling}. Distributional models for thematic fit estimation have been proposed by several authors \citep{Erk2007ASS,Baroni:2010:DMG:1945043.1945049,Erk2010AFC,lenci2011composing,Sayeed2015AnEO,Greenberg2015ImprovingUV,santus2017measuring,Tilk2016EventPM,hong2018learning}. 
While thematic fit datasets typically include human-elicited typicality scores for argument-filler pairs taken in isolation, DTFit includes tuples of arguments of different length, so that the typicality value of an argument depends on its interaction with the other arguments in the tuple. This makes it possible to model the dynamic aspect of argument typicality, since the expectations on an argument are dynamically updated as the other roles in the sentence are filled. The argument combinations in DTFit describe events associated with crowdsourced scores ranging from 1 (very atypical) to 7 (very typical). The dataset items are grouped into typical and atypical pairs that differ only for one argument, and divided into three subsets:

\begin{itemize}
\item 795 triplets, each differing only for the \textbf{Patient} role:
\begin{itemize}
\item \emph{sergeant}\_N \emph{assign}\_V \emph{mission}\_N (typical)
\item \emph{sergeant}\_N \emph{assign}\_V \emph{homework}\_N (atypical)
\end{itemize}
\item 300 quadruples, each differing only for the \textbf{Location} role:
\begin{itemize}
\item \emph{policeman}\_N \emph{check}\_V \emph{bag}\_N \emph{airport}\_N (typical)
\item \emph{policeman}\_N \emph{check}\_V \emph{bag}\_N \emph{kitchen}\_N (atypical)
\end{itemize}
\item 200 quadruples, each differing only for the \textbf{Instrument} role:
\begin{itemize}
\item \emph{painter}\_N \emph{decorate}\_V \emph{wall}\_N \emph{brush}\_N (typical)
\item \emph{painter}\_N \emph{decorate}\_V \emph{wall}\_N \emph{scalpel}\_N (atypical)
\end{itemize} 
\end{itemize}

\noindent{}However, the Instrument subset of DTFit was excluded from our current evaluation. After applying the threshold of $5$ for storing events in the \textsc{deg} (cf. Section \ref{sec:expGEK}), we found that the SDM coverage on this subset was too low. 

For each tuple in the DTFit dataset, the task for our models is to predict the upcoming argument on the basis of the previous ones. Given a model, we build a compositional vector representation for each dataset item by excluding the last argument in the tuple, and then we measured the cosine similarity between the resulting vector and the argument vector. Models are evaluated in terms of the Spearman correlation between the similarity scores and the human ratings.

As suggested by the experimental results of \cite{bicknell2010effects} and \cite{matsuki2011event}, the typicality of the described events has important processing consequences: atypical events lead to longer reading times and stronger N400 components, while typical ones are easier to process thanks to the contribution of \textsc{gek}. Thus, the task of modeling typicality judgements can be seen as closely related to modeling semantic processing complexity.

\subsection{Models Settings}

In this study, we compare the performance of SDM with three baselines. The simple additive model formulated in \cite{mitchell2010composition}, a smoothed additive model,  and a multi-layer Long-Short-Term-Memory (LSTM) neural language model trained against one-hot targets \citep{zaremba2014recurrent}.

The additive models \citep{mitchell2010composition} have been evaluated on different types of word embeddings. We compared their performances with SDM.\footnote{We also tested pointwise multiplicative models, but in our tasks the performances were extremely low, so they were omitted.} Despite their simplicity, previous evaluation studies on several benchmarks showed that such models can be difficult to beat, even for sophisticated compositionality frameworks \citep{rimell2016relpron,Arora:etal:2017,Tian:etal:2017}.

The embeddings we used in our tests are the \textsc{word2vec} models by \citet{mikolov2013distributed}, that is the Skip-Gram with Negative Sampling (\textbf{SG}) and the Continuous-Bag-of-Words (\textbf{CBOW}), and the \textbf{C-Phrase} model by \citet{kruszewski2015jointly}. The latter model incorporates information about syntactic constituents, as the principles of the model training are i.) to group the words together according to the syntactic structure of the sentences and ii.) to optimize simultaneously the context predictions at different levels of the syntactic hierarchy (e.g., given the training sentence \textit{A sad dog is howling in the park}, the context prediction will be optimized for \textit{dog, a dog, a sad dog} etc., that is for all the words that form a syntactic constituent). The performance of C-Phrase is particularly useful to assess the benefits of using vectors that encode directly structural/syntactic information.

We used the same corpora both for training the embeddings and for extracting the syntactic relations for \textsc{deg}. The training data come from the concatenation of three dependency-parsed corpora: the BNC \citep{leech1992100}, the Ukwac \citep{Baroni2009} and a 2018 dump of the English Wikipedia, for a combined size of approximately 4 billion tokens. The corpora were parsed with Stanford CoreNLP \citep{manning2014stanford}.
The hyperparameters of the embeddings were the following for all models: 400 dimensions, a context window of size 10, 10 negative samples, 100 as the minimum word frequency.\footnote{We tested different values for the dimension hyperparameter, and we noticed that vectors with higher dimensionality lead to constant improvements on the thematic fit datasets. The best results were obtained with 400 dimensions.}

\subsubsection{Simple Additive Models}
Our additive models, corresponding to a \textsc{sr} consisting of the $\overrightarrow{LC}$ component only, represent the meaning of a sentence $sent$ by summing the embeddings of its words:

\begin{equation}
\overrightarrow{sent} = \sum_{w \in sent}{\vec{w}}
\end{equation}

\noindent{}The similarity with the targets is measured with the cosine between the target vector and the sentence vector.

\subsubsection{Smoothed Additive Models}
\label{new_base}

These models are a smoothed version of the additive baseline, in which the final representation is simply the sum of the vectors of the words in the sentence, plus the top $k=5$ nearest neighbor of each word in the sentence.\footnote{We have experimented with $k=2,5,10$ and, although the scores do not significantly differ, this baseline model reports slightly better scores for $k=5$.}
Therefore, the meaning of a sentence $sent$ is obtained by:

\begin{equation}
\overrightarrow{sent} = \sum_{w \in sent}{\left(\vec{w} + \sum_{x \in N_5(w)} \vec{x}\right)}
\end{equation}

\noindent{}where $N_k(w)$ is the set of the $k$ nearest neighbors of $w$.
Compared to the \textsc{gek} models, the smoothed additive baseline modifies the sentence vector by adding the vectors of related words. Thus, it represents a useful comparison term for understanding the actual added value of the structural aspects of SDM.\footnote{We would like to thank one of the anonymous reviewers for the suggestion.}

\subsubsection{The Structured Distributional Models}
\label{sec:expGEK}

The SDM introduced in Section \ref{sec:Model} consists of a full \textsc{sr} including the linguistic conditions vector $\overrightarrow{LC}$ and the event knowledge vector $\overrightarrow{AC}$. In this section, we detail the hyperparameter setting for the actual implementation of the model.

\paragraph{\textbf{Distributional Event Graph}}

We included in the graph only events with a minimum frequency of 5 in the training corpora. The edges of the graph were weighted with \emph{Smoothed LMI}.
Given a triple composed by the words $w_1$ and $w_2$, and a syntactic relation $s$ linking them, we computed its weight by using a smoothed version of the Local Mutual Information \citep{Evert2004TheSO}:

\begin{equation}
LMI_\alpha(w_1, w_2, s) = f(w_1, w_2, s) * log(\frac{P(w_1, w_2, s)}{P(w_1)*P_\alpha(w_2)*P(s)})
\end{equation}\\

\noindent{}where the smoothed probabilities are defined as follows:

\begin{equation}
P_\alpha(x) = \frac{f(x)^\alpha}{\sum_x{f(x)^\alpha}}
\end{equation}\\

\noindent{}This type of smoothing, with $\alpha=0.75$, was chosen to mitigate the bias of MI statistical association measures towards rare events \citep{levy2015improving}. While this formula only involves pairs (as only pairs were employed in the experiments), it is easily extensible to more complex tuples of elements.

\paragraph{\textbf{Re-ranking settings}}
For each word in the dataset items, the top 50 associated words were retrieved from \textsc{deg}. Both for the re-ranking phase and for the construction of the final representation, the event knowledge vectors (i.e., the role vectors $\overrightarrow{r}$ and the \textsc{ac} vector \overrightarrow{AC}) are built from the top 20 elements of each weighted list. As detailed in Section \ref{sec:MCF}, the ranking process in SDM can be performed in the forward direction and in the backward direction at the same time (i.e., the \textsc{ac} can be used to re-rank newly retrieved information and vice versa, respectively), but for simplicity we only implemented the forward ranking. 

\paragraph{\textbf{Scoring}}
As in SDM the similarity computations with the target words involves two separate vectors, we combined the similarity scores with addition. Thus, given a $target$ word in a sentence $sent$, the score for SDM will be computed as:

\begin{equation}
score(target, sent) = cos(\overrightarrow{target}, \overrightarrow{LC}(sent)) + cos(\overrightarrow{target}, \overrightarrow{AC}(sent))
\end{equation}\\

In all settings, we assume the model to be aware of the syntactic parse of the test items. In DTFit, word order fully determines the syntactic constituents, as the sentences are always in the \textit{subject verb object [location-obl|instrument-obl]} order. In RELPRON, on the other hand, the item contains information about the relation that is being tested: in the \textit{subject} relative clauses, the properties always show the \textit{verb} followed by the \textit{argument} (e.g., \textit{telescope: device that detects planets}), while in the \textit{object} relative clauses the properties always present the opposite situation (e.g., \textit{telescope: device that observatory has}). In the present experiments, we did not use the predictions on non-expressed arguments to compute $\overrightarrow{AC}$, and we restricted the evaluation to the representation of the target argument. For example, in the DTFit Patients set, $\overrightarrow{AC}(sent)$ only contains the $\overrightarrow{dobj}$ centroid.

\subsubsection{LSTM Neural Language Model.} 
We also compared the additive vector baselines and SDM with an LSTM neural network, taking as input \textsc{word2Vec} embeddings. For every task, we trained the LSTM on syntactically-labeled tuples (extracted from the same training corpora used for the other models), with the objective of predicting the relevant target. In DTFit, for example, for the Location task, in the tuple \textit{student learn history library}, the network is trained to predict the argument \textit{library} given the tuple \textit{student learn history}. Similarly, in RELPRON, for the tuple \textit{engineer patent design}, the LSTM is trained to predict \textit{engineer} in the subject task and \textit{design} in the object task, given \textit{patent design} and \textit{engineer patent} respectively. 

In both DTFit and RELPRON, for each input tuple, we took the top $N$ network predictions (we tested with $N={3, 5, 10}$, and we always obtained the best results with $N = 10$), we averaged their respective word embeddings, and we used the vector cosine between the resulting vector and the embedding of the target reported in the gold standard.

The LSTM is composed by: i.) an input layer of the same size of the \textsc{word2Vec} embeddings (400 dimensions, with dropout=0.1); ii.) a single-layer monodirectional LSTM with $l$ hidden layers (where $l=2$ when predicting Patients and $l=3$ when predicting Locations) of the same size of the embeddings; iii.) a linear layer (again with dropout= 0.1) of the same size of the embeddings, which takes in input the average of the hidden layers of the LSTM; iv.) and finally a softmax layer that projects the filler probability distribution over the vocabulary.

\section{Results and Discussion}
\label{sec:Res}

\subsection{RELPRON}
\label{sec:REL}

Given the targets and the composed vectors of all the definitions in RELPRON, we assessed the cosine similarity of each pair and computed the Mean Average Precision scores shown in Table \ref{addgek}. First of all, the Skip-Gram based models always turn out to be the best performing ones, with rare exceptions, closely followed by the C-Phrase ones. The scores of the additive models are slightly inferior, but very close to those reported by \citet{rimell2016relpron}, while the LSTM model lags behind vector addition, improving only when the parameter $N$ is increased. Results seem to confirm the original findings: even with very complex models (in that case, the Lexical Function Model by Paperno et al. 2014), it is difficult to outperform simple vector addition in compositionality tasks. 

\begin{table}[t]
\begin{tabular}{l|llcccc}
\multicolumn{1}{c}{} & \multicolumn{1}{c}{} & \multicolumn{1}{l}{\textbf{Word combination}}    & \textbf{R\&al.} & \textbf{SG}    & \textbf{CBOW}   & \textbf{C-Phrase} \\ \hline
\multirow{6}{*}{{\centering \textbf{Additive}}}
& & verb               & 0.18          & 0.16     & 0.16    & 0.13    \\
& & arg                         & 0.35          & 0.33    & 0.32    & 0.37    \\
& & head noun+verb              & 0.26          & 0.26    & 0.25    & 0.21    \\
& & head noun+arg               & 0.45          & 0.44    & 0.46    & 0.45    \\
& & verb+arg                    & 0.40          & 0.43    & 0.36    & 0.41    \\
& & head noun+verb+arg          & \textbf{0.50}          & \textbf{0.50}    & 0.47   & 0.47   \\ \hline
\multirow{6}{*}{{\centering \textbf{Smoothed}}}
& & verb         &        -                & 0.15     & 0.16    & 0.14    \\
& & arg           &          -              & 0.35    & 0.33    & 0.40    \\
& & head noun+verb      &     -             & 0.24    & 0.23    & 0.22    \\
& & head noun+arg        &       -          & 0.45    & 0.46    & 0.49    \\
& & verb+arg              &      -          & 0.41    & 0.36    & 0.41    \\
& & head noun+verb+arg     &      -         & \textbf{0.49}    & 0.46   & 0.47   \\ \hline
\multirow{1}{*}{{\centering \textbf{LSTM}}}
%& & LSTM\_3 & - & -& 0.22& -\\
%& & LSTM\_5 & - & -& 0.26 & -\\
& & LSTM\_10 & - & 0.10 & 0.32& - \\
\hline 
\multirow{6}{*}{\textbf{SDM}}
& & verb & - & 0.21    & 0.20  & 0.19    \\
& & arg & - & 0.38    & 0.36   & 0.41    \\
& & head noun+verb & - & 0.27    & 0.28   & 0.26    \\
& & head noun+arg & - & 0.50    & 0.50   & 0.50    \\
& & verb + arg & - & 0.41    & 0.36   & 0.41    \\
& & head noun + verb + arg & - & \textbf{0.54}    & 0.52   & \textbf{0.54}  \\ \hline
\end{tabular}
\caption{Results for the Vector Addition Baseline, Smoothed Vector Addition Baseline, LSTM and the Structured Distributional Model (SDM) on the RELPRON development set (Mean Average Precision scores). Rows refer to the different word combinations tested in Rimell et al. 2016 (R\&al.).}
\label{addgek}
\end{table}

Interestingly, SDM shows a constant improvement over the simple vector addition equivalents (Table \ref{addgek}), with the only exception of the composition of the verb and the argument. All the results for the $headNoun + verb + arg$ composition are, to the best of our knowledge, the best scores reported so far on the dataset. 
Unfortunately, given the relatively small size of RELPRON, the improvement of the \textsc{gek} models fails to reach significance ($p > 0.1$ for all comparisons between a basic additive model and its respective augmentation with \textsc{deg}, $p-$values computed with the Wilcoxon rank sum test). Compared to SDM, the Smoothed Vector Addition baseline seems to be way less consistent (Table \ref{addgek}): for some combinations and for some vector types, adding the nearest neighbors is detrimental. We take these results as supporting the added value of the structured event knowledge and the \textsc{sr} update process in SDM, over the simple enrichment of vector addition with nearest neighbors.
Finally, we can notice that the Skip-Gram vectors have again an edge over the competitors, even over the syntactically-informed C-Phrase vectors.

\begin{table}[t]
\begin{tabular}{l|lcccc}
\multicolumn{1}{c}{} & \multicolumn{1}{c}{} &  \textbf{Dataset} & \textbf{SG} & \textbf{CBOW} & \textbf{C-Phrase} \\ \hline
\multirow{2}{*}{{\centering \textbf{Additive}}}
& & Patients & \textbf{0.63} & 0.52 & 0.60   \\
& & Locations   & \textbf{0.74}  & 0.70  & 0.74 \\ \hline
\multirow{2}{*}{{\centering \textbf{Smoothed}}}
& & Patients   & \textbf{0.58} & 0.51 & \textbf{0.58}   \\
& & Locations    & 0.74 & 0.71  & \textbf{0.76} \\ \hline
\multirow{2}{*}{{\centering \textbf{LSTM}}}
& & Patients   & ns & 0.42 & -   \\
& & Locations    & 0.58 & \textbf{0.60}  & - \\ \hline
\multirow{2}{*}{{\centering \textbf{SDM}}}
& & Patients      & 0.65 & 0.62** & \textbf{0.66} *  \\
& & Locations     & 0.75 & 0.74  & \textbf{0.76} \\ \hline
\end{tabular}
\caption{Results for the Vector Addition Baseline, Smoothed Vector Addition Baseline, LSTM and the Structured Distributional Model (SDM) on the Patients and Locations subsets of DTFit. The scores are expressed in terms of Spearman correlation with the gold standard ratings. The LSTM scores refer to the best configuration, with $N = 10$ and vectors of size 400. The statistical significance of the improvements over the additive baseline is reported as follows: * $p < 0.05$, ** $p < 0.01$ (p-values computed with Fisher's r-to-z transformation, one-tailed test). ns = non significant correlation.}
\label{tab:dtfit}
\end{table}

\subsection{DTFit}
\label{sec:dtfit}

At a first glance, the results on DTFit follow a similar pattern (Table \ref{tab:dtfit}): The three embedding types perform similarly, although in this case the CBOW vectors perform much worse than the others in the Patients dataset. LSTM also largely lags behind all the additive models, showing that thematic fit modeling is not a trivial task for language models, and that more complex neural architectures are required in order to obtain state-of-the-art results \citep{Tilk2016EventPM}.\footnote{It should also be noticed that our LSTM baseline has been trained on simple syntactic dependencies, while state-of-the-art neural models rely simultaneously on dependencies and semantic role labels \citep{Tilk2016EventPM,hong2018learning}.} 

The results for SDM again show that including the \textsc{deg} information leads to improvements in the performances (Table \ref{tab:dtfit}). While on the Locations the difference is only marginal, also due to the smaller number of test items, two models out of three showed significantly higher correlations than their respective additive baselines. The increase is particularly noticeable for the CBOW vectors that, in their augmented version, manage to fill the gap with the other models and to achieve a competitive performance. However, it should also be noticed that there is a striking difference between the two subsets of DTFit: while on patients the advantage of the \textsc{gek} models on both the baselines is clear, on locations the results are almost indistinguishable from those of the smoothed additive baseline, which simply adds the nearest neighbours to the vectors of the words in the sentence. This complies with previous studies on thematic fit modeling with dependency-based distributional models \citep{Sayeed2015AnEO,santus2017measuring}. Because of the ambiguous nature of the prepositions used to identify potential locations, the role vectors used by SDM can be very noisy. Moreover, since most locative complements are optional adjuncts, it is likely that the event knowledge extracted from corpora contain a much smaller number of locations. 
Therefore, the structural information about locations in \textsc{deg} is probably less reliable and does not provide any clear advantage compared to additive models.

Concerning the comparison between the different types of embeddings, Skip-Gram still retains an advantage over C-Phrase in its basic version, while it is outperformed when the latter vectors are used in SDM. However, the differences are clearly minimal,
suggesting that the structured knowledge encoded in the C-Phrase embeddings is not a plus for the thematic fit task. Concerning this point, it must be mentioned that most of the current models for thematic fit estimation rely on vectors relying either on syntactic information \citep{Baroni:2010:DMG:1945043.1945049,Greenberg2015ImprovingUV,santus2017measuring,chersoni2017structure} or semantic roles \citep{Sayeed2015AnEO,Tilk2016EventPM}. On the other hand, our results comply with studies like \citet{lapesa2017large}, who reported comparable performance for bag-of-words and dependency-based models on several semantic modeling tasks, thus questioning whether the injection of linguistic structure in the word vectors is actually worth its processing cost.
However, this is the first time that such a comparison is carried out on the basis of the DTFit dataset, while previous studies proposed slightly different versions of the task and evaluated their systems on different benchmarks.\footnote{In datasets such as \citet{mcrae1998modeling} and \citet{pado2007integration}, the verb-filler compatibility is modeled without taking into account the influence of the other fillers. On the other hand, studies on the composition and update of argument expectations generally propose evaluations in terms of classification tasks \citep{lenci2011composing,chersoni2017structure} instead of assessing directly the correlation with human judgements.} A more extensive and in-depth study is required in order to formulate more conclusive arguments on this issue. 

Another constant finding of previous studies on thematic fit modeling was that high-dimensional, count-based vector representations perform generally better than dense word embeddings, to the point that \citet{Sayeed2016ThematicFE} stressed the sensitivity of this task to linguistic detail and to the interpretability of the vector space. Therefore, we tested whether vector dimensionality had an impact on task performance (Table \ref{models}). Although the observed differences are generally small, we noticed that higher-dimensional vectors are generally better in the DTFit evaluation and, in one case, the differences reach a marginal significance (i.e., the difference between the 100-dimensional and the 400-dimensional basic Skip-Gram model is marginally significant at $p<0.1$). This point will also deserve future investigation, but it seems plausible that for this task embeddings benefit from higher dimensionality for encoding more information, as it has been suggested by Sayeed and colleagues. However, these advantages do not seem to be related to the injection of linguistic structure directly in the embeddings (i.e., not to the direct use of syntactic contexts for training the vectors), as bag-of-words models perform similarly to - if not better than - a syntactic-based model like C-Phrase. We leave to future research a systematic comparison with sparse count-based models to assess whether interpretable dimensions are advantageous for modeling context-sensitive thematic fit.

\subsection{Error Analysis}

One of our basic assumptions about \textsc{gek} is that semantic memory stores representations of typical events and their participants. Therefore, we expect that integrating \textsc{gek} into our models might lead to an improvement especially on the typical items of the DTFit dataset. A quick test with the correlations revealed that this is actually the case (Table \ref{typ}): all models showed increased Spearman correlations on the tuples in the typical condition (and in the larger Patients subset of DTFit, the increase is significant at $p < 0.05$ for the CBOW model), while they remain unchanged or even decrease for the tuples in the atypical conditions. Notice that this is true only for SDM, which is enriched with \textsc{gek}. On the other hand, the simple addition of the nearest neighbors never leads to improvements, as proved by the low correlation scores of the smoothed additive baseline. As new and larger datasets for compositionality tasks are currently under construction \citep{vassallo2018event}, it will be interesting to assess the consistency of these results.

\begin{table}[t]
\begin{tabular}{ccc}
\textbf{Dimensions} & \textbf{Additive}    & \textbf{SDM} \\ \hline
100        & 0.58 & 0.63    \\
200        & 0.58      &    0.63     \\
300        & 0.60 & 0.64   \\
400        & 0.64 & \textbf{0.65}\\  \hline
\end{tabular}
\caption{Spearman correlations of the Vector Addition Baseline and the Structured Distributional Model (SDM) based on Skip-Gram on the DTFit patients subset.}
\label{models}
\end{table}

\begin{table}[t]
\begin{tabular}{lcccl}
\textbf{Model}   & \textbf{Additive} &  \textbf{Smoothed} & \textbf{SDM} & \textbf{$\Delta$} \\ 
\hline
CBOW    & 0.18    &  0.18    & 0.30      & + 0.12 *     \\
SG      & 0.29      &  0.24        & 0.33   & + 0.04     \\
C-Phrase & 0.30     & 0.29        & \textbf{0.37}    & + 0.07 \\\hline
\end{tabular}
\caption{Comparison of the performance of the Vector Addition Baseline, Smoothed Vector Addition Baseline, and the Structured Distributional Model (SDM) on the typical items of DTFit Patients (Spearman correlations). $\Delta$ reports the SDM improvements over the basic additive models. Significance is noted with the following notation: * $p < 0.05$.}
\label{typ}
\end{table}

Turning to the RELPRON dataset, we noticed that the difference between subject and object relative clauses is particularly relevant for SDM, which generally shows better performances on the latter. Table \ref{tab:verbarg} summarizes the scores component on the two subsets. While relying on syntactic dependencies, SDM also processes properties in linear order: the \textit{verb}$+$\textit{arg} model, therefore, works differently when applied to \textit{subject} clauses than to \textit{object} clauses. In the \textit{subject} case, in fact, the verb is found first, and then its expectations are used to re-rank the object ones. In the \textit{object} case, on the other hand, things proceed the opposite way: at first the subject is found, and then its expectations are used to re-rank the verb ones. Therefore, the event knowledge triggered by the verb seems not only less informative than the one triggered by the argument, but it is often detrimental to the composition process.

\begin{table}[t]
\begin{tabular}{lcc|cc|cc}
     \textbf{SDM}       & \multicolumn{2}{c}{\textbf{SG}} & \multicolumn{2}{c}{\textbf{CBOW}} & \multicolumn{2}{c}{\textbf{C-Phrase}} \\ \hline
         \textbf{Subset}   & \textit{sbj}   & \textit{obj}       & \textit{sbj} & \textit{obj}& \textit{sbj}& \textit{obj} \\ \hline
%verb        & 0.22       & 0.22      & 0.23        & 0.20       & 0.26         & 0.16         \\
%arg         & 0.45       & 0.41      & 0.44        & 0.39       & 0.48         & 0.45         \\
head noun+verb     & 0.29       & 0.31      & 0.32        & 0.32       & 0.29         & 0.28         \\
head noun+arg      & 0.54       & 0.57      & 0.54        & 0.56       & 0.56         & 0.57         \\
verb+arg    & 0.45       & 0.47      & 0.40        & 0.43       & 0.47         & 0.47         \\
head noun+verb+arg & 0.56       & 0.61      & 0.58        & 0.57       & 0.60         & 0.58 \\ \hline
\end{tabular}
\caption{Comparison of the Structured Distributional Model (SDM) performance (MAP) on the subject and object relative clauses in RELPRON.}
\label{tab:verbarg}
\end{table}

\section{Conclusion}
\label{sec:conc}

In this contribution, we introduced a Structured Distributional Model (SDM) that represents sentence meaning with formal structures derived from DRT and including embeddings enriched with event knowledge. This is modeled with a Distributional Event Graph that represents events and their prototypical participants with distributional vectors linked in a network of syntagmatic relations extracted from parsed corpora. The compositional  construction of sentence meaning in SDM is directly inspired by the principles of dynamic semantics. Word embeddings are integrated in a dynamic process to construct the semantic representations of sentences: contextual event knowledge affects the interpretation of following expressions, which cue new information that updates the current context. 

Current methods for representing sentence meaning generally lack information about typical events and situation, while SDM rests on the assumption that such information can lead to better compositional representations and to an increased capacity of modeling typicality, which is one striking capacity of the human processing system. This corresponds to the hypothesis by \citet{baggio2011balance} that semantic compositionality actually results from a balance between storage and computation: on the one hand, language speakers rely on a wide amount of stored events and scenarios for common, familiar situations; on the other hand, a compositional mechanism is needed to account for our understanding of new and unheard sentences. Processing complexity, as revealed by effects such as the reduced amplitude of the N400 component in ERP experiments, is inversely proportional to the typicality of the described events and situations: The more they are typical, the more they will be coherent with already-stored representations.

We evaluated SDM on two tasks, namely a classical similarity estimation tasks on the target-definition pairs of the RELPRON dataset \citep{rimell2016relpron} and a thematic fit modeling task on the event tuples of the DTFit dataset \citep{vassallo2018event}. Our results still proved that additive models are quite efficient for compositionality tasks, and that integrating the event information activated by lexical items improves the performance on both the evaluation datasets. Particularly interesting for our evaluation was the performance on the DTFit dataset, since this dataset has been especially created with the purpose of testing computational models on their capacity to account for human typicality judgments about event participants.
The reported scores on the latter dataset showed that not only SDM improves over simple and smoothed additive models, but also that the increase in correlation concerns the dataset items rated as most typical by human subjects, fulfilling our initial predictions. 

Differently from other distributional semantic models tested on the thematic fit task, `structure' is now externally encoded in a graph, whose nodes are embeddings, and not directly in the dimension of the embeddings themselves. The fact that the best performing word embeddings in our framework are the Skip-Gram ones is somewhat surprising, and against the finding of previous literature in which bag-of-words models were always described as struggling on this task \citep{baroni2014don,Sayeed2016ThematicFE}. Given our results, we also suggested that the dimensionality of the embeddings could be an important factor, much more than the choice of training them on syntactic contexts.

\bibliographystyle{chicago-nle}
\bibliography{bibliography}

\label{lastpage}

\end{document}